\documentclass[10pt,twocolumn,letterpaper]{article}

\usepackage[pagenumbers]{iccv} 

\definecolor{iccvblue}{rgb}{0.21,0.49,0.74}
\usepackage[pagebackref,breaklinks,colorlinks,allcolors=iccvblue]{hyperref}

\def\paperID{} 
\def\confName{ICCV}
\def\confYear{2025}

\title{T-3DGS: Removing Transient Objects for 3D Scene Reconstruction}

\author{
Alexander Markin$^{1}$\thanks{Equal contribution.} \quad
Vadim Pryadilshchikov$^{1}$\footnotemark[1] \quad
Artem Komarichev$^{1}$ \quad
Ruslan Rakhimov$^{2}$ \\
Peter Wonka$^{3}$ \quad
Evgeny Burnaev$^{1,4}$ \\
$^1$Skoltech, Russia \quad
$^2$T-Tech, Russia \quad
$^3$KAUST, Saudi Arabia \quad
$^4$AIRI, Russia
}

\begin{document}

\twocolumn[{%
\renewcommand\twocolumn[1][]{#1}%
\maketitle
\begin{center}
    \centering
    \includegraphics[width=.99\textwidth]{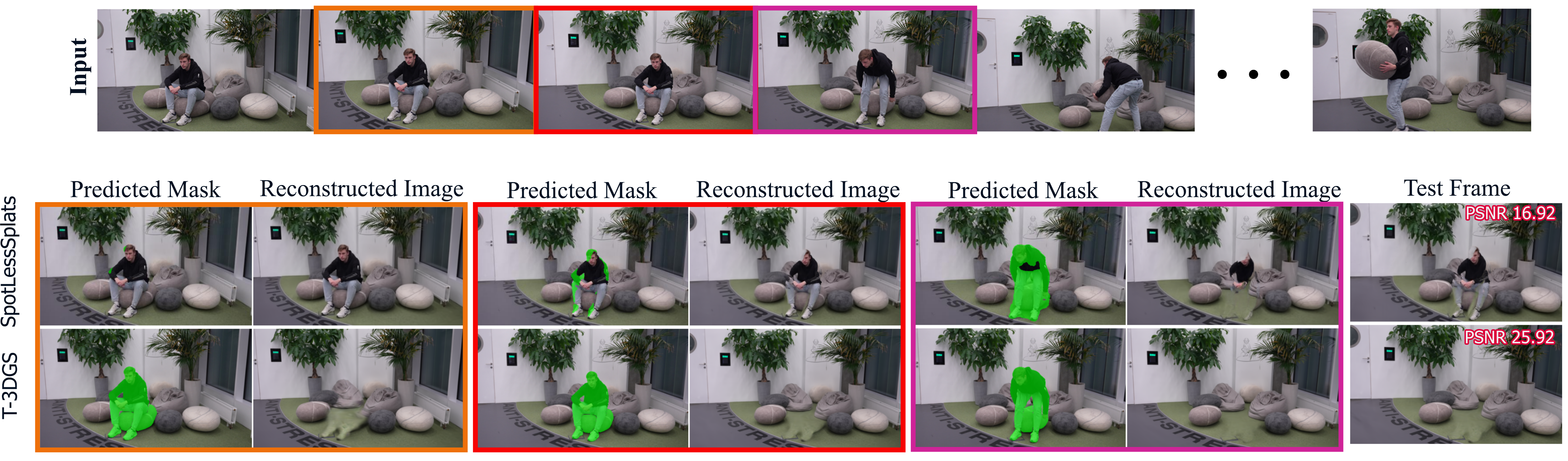}
    \captionsetup{type=figure}
    \caption{Existing state-of-the-art methods, such as SpotLessSplats~\cite{sabour2024spotlesssplats}, often struggle to correctly identify transient and semi-transient objects, leading to artifacts in 3D scene reconstruction. Our proposed \emph{T-3DGS} method accurately detects all transient distractors, generates clean masks, and propagates them across frames. By effectively masking transient objects, \emph{T-3DGS} enables high-fidelity novel view synthesis and significantly improves reconstruction quality from real-world image and video sequences.}
    \label{fig:teaser}
\end{center}
}]
\footnotetext[1]{Equal contribution.}
\footnotetext[2]{Correspondence to: a.komarichev@skoltech.ru.}

\maketitle
\begin{abstract}
Transient objects in video sequences can significantly degrade the quality of 3D scene reconstructions. To address this challenge, we propose T-3DGS, a novel framework that robustly filters out transient distractors during 3D reconstruction using Gaussian Splatting. Our framework consists of two steps. First, we employ an unsupervised classification network that distinguishes transient objects from static scene elements by leveraging their distinct training dynamics within the reconstruction process. Second, we refine these initial detections by integrating an off-the-shelf segmentation method with a bidirectional tracking module, which together enhance boundary accuracy and temporal coherence. Evaluations on both sparsely and densely captured video datasets demonstrate that T-3DGS significantly outperforms state-of-the-art approaches, enabling high-fidelity 3D reconstructions in challenging, real-world scenarios.  More results and code are available at \url{https://transient-3dgs.github.io/}
\end{abstract}

\vspace{-0.7cm}
\section{Introduction}

Novel view synthesis and 3D scene reconstruction from multiple 2D images or videos are critical, rapidly evolving areas in computer vision. Neural Radiance Fields (NeRF)~\cite{mildenhall2021nerf} and 3D Gaussian Splatting (3DGS)~\cite{kerbl20233d} have shown remarkable improvements in novel view synthesis on complex scenes.
NeRF implicitly represents the scene as a volumetric function, and
3DGS explicitly represents it as a set of 3D Gaussians. Both approaches produce high-quality realistic images. There are multiple follow-up works for diverse downstream applications, including 3D scene reconstruction~\cite{wang2021neus,li2023neuralangelo,guedon2024sugar},
3D synthesis~\cite{poole2022dreamfusion,wynn2023diffusionerf,tang2025lgm},
semantic and language integration into 3D representations~\cite{siddiqui2023panoptic,kerr2023lerf,shi2024language}.

Both 3D Gaussian Splatting and NeRF optimize 3D scene reconstruction using photometric losses. High-quality results are achieved under the assumption that the captured scene is completely static and does not include any \emph{distractors}, such as moving objects (i.e. transient objects), shadows, lightning changes, etc. In real-world scenarios, this assumption can hardly be satisfied. Even when carefully captured, recordings often contain moving people, cars, or other dynamic objects with their shadows, especially in locations that are tourist landmarks. Ignoring distractors during scene optimization results in undesired blurring effects and floating artifacts. At the same time, removing such distractors from the captured recordings is challenging and limits the widespread usage of NeRF and 3DGS. Manually annotating distractors is labor intensive. Another approach is to utilize pre-trained segmentation models to locate transient distractors. This approach has two main limitations: 1) it needs prior knowledge of transients as a semantic class, and 2), more importantly, existing segmentation models cannot distinguish between static and dynamic objects of the same semantic class.
Additionally, we would like to identify semi-transient objects in recordings and remove them from the scene. We define a semi-transient object as an object that has both dynamic and static states during the capturing process, e.g. a pushed chair stops after some time and becomes a fully static object. Therefore, we need more robust identification methods for transient and semi-transient distractors throughout the captured recordings.

We introduce T-3DGS, a novel approach for 3D static scene reconstruction from monocular video in uncontrolled settings. Our method includes an unsupervised transient detector and a transient mask propagation framework. Relying solely on image residuals for transient identification is unreliable due to issues such as appearance changes and color similarity to the background~\cite{ren2024nerf,sabour2024spotlesssplats}. To address this issue, we develop a divergence-based technique on top of the uncertainty modeling approach~\cite{kulhanek2024wildgaussians} to detect transients. It helps improve mask accuracy and significantly reduce misclassifications of transient objects.

Our experiments show that concurrent works~\cite{sabour2024spotlesssplats,ungermann2024robust} fail to remove semi-transient distractors (Fig.~\ref{fig:teaser}). We introduce a mask propagation framework for extracting object-aware masks that improve consistency in case of semi-transient distractors. Our method remains robust to all types of distractors. Additionally, we present the novel \emph{T-3DGS dataset} with challenging scenes featuring semi-transient and slow-moving objects. Evaluations on both casual scenes~\cite{ren2024nerf,sabour2023robustnerf} and our dataset show our method outperforms state-of-the-art approaches in reconstruction quality.

Our key contributions, which together ensure consistent detection and removal of transient objects for improved 3D reconstruction, include:
\begin{itemize}
    \item Generalized uncertainty modeling for efficient transient object identification;
    \item A divergence-based approach that leverages semantic consistency between reference and reconstructed frames for identifying transient objects;
    \item A robust video object segmentation module that tracks objects across varying frame rates and semi-transient behaviors;
    \vspace{10pt}
    \item A challenging new dataset featuring diverse scenes with semi-transient distractors and slow-moving objects;
    \item State-of-the-art performance on benchmark datasets for robust static scene reconstruction.
\end{itemize}

\section{Related Work}
\label{sec:relatedwork}

We provide a brief review of the works on Neural Radiance Fields and 3D Gaussian Splatting with a focus on removing non-static distractors in the scene.

Neural Radiance Fields (NeRFs)~\cite{mildenhall2021nerf} are widely adopted methods for high-quality scene reconstruction and novel view synthesis of 3D scenes. The seminal work 3D Gaussian Splatting~\cite{kerbl20233d} employs Gaussian primitives to model scenes instead of relying on continuous volumetric representations. This method has recently gained popularity as a faster alternative to NeRFs.

\noindent\textbf{Handling Distractors in NeRFs.} NeRF-W~\cite{martin2021nerf} and RobustNeRF~\cite{sabour2023robustnerf} are two pioneering works approaching the problem in a similar way.
NeRF-W reconstructs both static background and transients combined with a data-dependent uncertainty field. RobustNeRF utilizes Iteratively Reweighted Least Squares for transient object identification and removal. Both methods rely on color residual supervision and frequently misclassify transient objects and backgrounds that share similar colors. Additionally, they both require careful hyper-parameters tuning. NeRF On-the-go~\cite{ren2024nerf} utilizes DINOv2 features~\cite{oquab2023dinov2} to identify and eliminate distractors by predicting uncertainties through a shallow MLP and can deal with more complex scenes than RobustNeRF. NeRF-HuGS~\cite{chen2024nerf} utilizes two types of heuristics: 1) COLMAP-based~\cite{schonberger2016structure} features combined with SAM~\cite{kirillov2023segment} and 2) residual-based heuristics to identify and remove transient distractors. Their method lacks robustness to heavy transient distractions, as both heuristics are unstable under such conditions, as demonstrated in~\cite{ren2024nerf}.

\noindent\textbf{Extracting Features from Vision Foundation Models}. Vision Foundation Models (VFMs) are trained on large-scale data, enabling strong generalization to unseen domains or novel tasks. Task-agnostic models trained through self-distillation like DINO~\cite{caron2021emerging,oquab2023dinov2} learn features that can be generalized for multiple vision tasks.

\noindent\textbf{Video Object Segmentation}. The goal of semi-supervised VOS is to identify when an object appears for the first time and then track it throughout the video. Several recent approaches based on transformers~\cite{cheng2023tracking,cheng2024putting} have been proposed. However, current methods suffer from mask inconsistencies, particularly when objects disappear and reappear in the video. Additionally, these methods assume that the input has a high frame rate, and they become unstable when the frame rate is low. In our work, we address these shortcomings.

\begin{figure*}
  \centering
  \includegraphics[trim={4cm 5cm 4cm 4cm},clip,width=\textwidth]{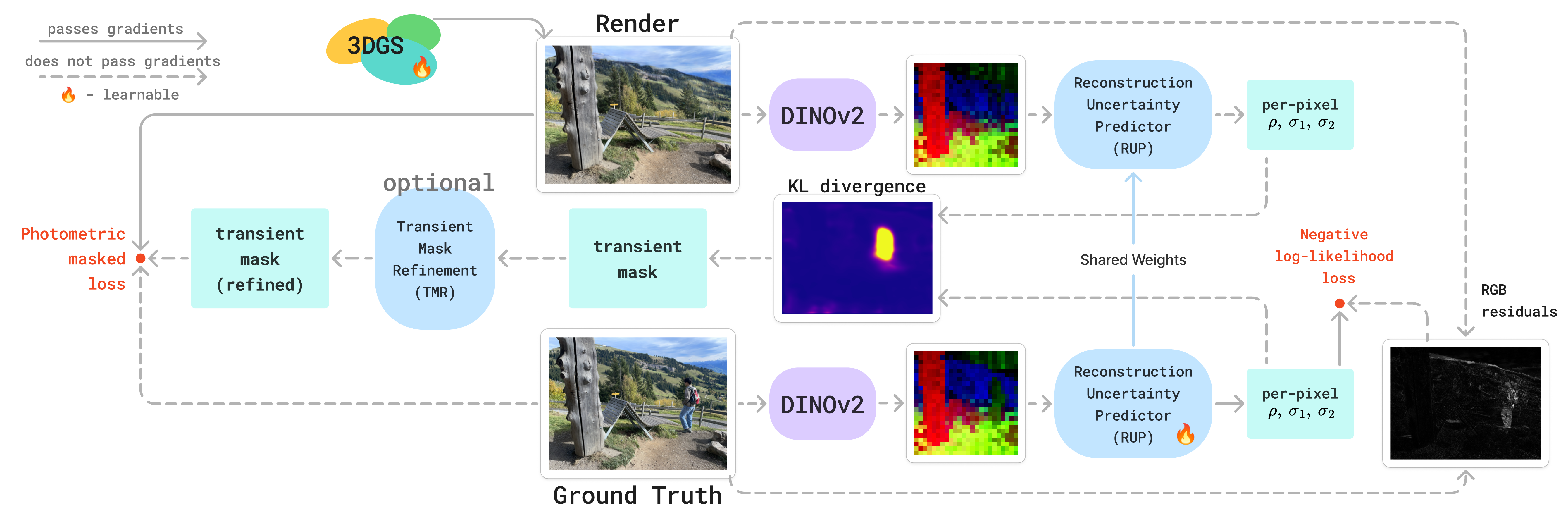}
  \vspace{-0.5cm}
  \caption{\emph{Overview of the Proposed T-3DGS Architecture.} We introduce a modified version of 3D Gaussian Splatting, incorporating a masked loss term $\mathcal{L}_{\text{masked}}$ as described in Eq.~\ref{masked_loss}. In each iteration, we start by rendering a reconstruction of a randomly sampled reference image. We compute residuals, along with DINOv2 features from both the ground truth and rendered images. These features are then fed to our \emph{RUP} model to predict the per-pixel covariance matrix for both images. We calculate binary masks based on the divergence of these distributions (as specified in Eq.~\ref{divergence_criterion}). Subsequently, we compute the likelihood as described in Eq.~\ref{likelihood} and update the parameters of the \emph{RUP} model via backpropagation, as indicated by the dashed lines. Additionaly, for some scenes, we incorporate a SAM-based mask refiner module (\emph{TMR}), which further enhances the consistency and sharpness of the masks.}
  \label{fig:pipeline}
\end{figure*}

\noindent\textbf{Handling Distractors in 3DGS.} Several works address the training of 3DGS on unconstrained, in-the-wild photo collections. SWAG~\cite{dahmani2024swag} improves robustness of 3DGS by learning an appearance embedding space and image-dependent opacity variations to handle transient objects better. Gaussians in the Wild (GS-W)~\cite{zhang2024gaussian} utilizes CNN features to capture dynamic and intrinsic appearances from a reference image. Wild-GS~\cite{xu2024wild} explicitly learns appearance embeddings by sampling the triplane from the reference image. Robust 3DGS~\cite{ungermann2024robust} proposes a self-supervised approach to identify transient distractors by utilizing image residuals and leveraging a pre-trained segmentation network to produce object-aware masks. SpotLessSplats~\cite{sabour2024spotlesssplats} proposes a method to identify transient objects by utilizing pre-computed feature maps from a foundation model~\cite{tang2023emergent} coupled with a robust optmization of 3DGS. These works~\cite{dahmani2024swag,zhang2024gaussian,xu2024wild,ungermann2024robust,sabour2024spotlesssplats} suffer from: 1) the need for hyper-parameter tuning, such as threshold parameters; 2) inaccurate prediction of transient masks across the video; and 3) reliance on image residuals, leading to the false detection of transients, as shown in Fig.~\ref{fig:teaser}. In our approach, we aim to address the limitations of the current works by identifying transients more accurately and consistently across video frames.

\section{Method}

We propose a novel approach to reconstructing static scenes from unconstrained videos that contain dynamic objects, utilizing 3D Gaussian Splatting (3DGS). Our method, illustrated in Fig.~\ref{fig:pipeline}, introduces two key components designed to handle dynamic objects effectively: (1) \textbf{reconstruction uncertainty predictor (\emph{RUP})}, and (2) \textbf{transient mask refiner (\emph{TMR})}. The transient area detection component, implemented through our transient mask learning predictor, identifies regions containing dynamic objects by predicting per-pixel probabilities using semantic features. The transient mask refiner improves transient detections in both spatial and temporal domains by leveraging SAM2~\cite{ravi2024sam} to propagate transient masks across multiple frames, facilitating artifact-free reconstruction.

\subsection{3D Gaussian Splatting}

Our method is based on 3DGS~\cite{kerbl20233d}. Given a set of posed images $\{I_n\}_{n=1}^N, I_n\in\mathbb{R}^{H\times W\times C}$, 3DGS represents a 3D scene as a set of anisotropic Gaussians $\{\mathcal{G}_{i}\}$, where each Gaussian is represented by its position (mean) $\mu_i$, a positive semi-definite covariance matrix $\Sigma_i$, an opacity $\alpha_i$, and a view-dependent appearance component (color) parametrized by spherical harmonics (SH)~\cite{ramamoorthi2001efficient}. 3DGS representation is learned through optimization of Gaussian parameters via stochastic gradient descent.

Each 3D Gaussian is projected onto the image plane through a differentiable rasterization process to render an image from a specific viewpoint. First, the 3D Gaussian's covariance matrix $\Sigma_i$ is projected to obtain a 2D covariance matrix $\Sigma'_i$ in screen space: $\Sigma'_i = JW\Sigma_i W^TJ^T$,
where $W$ is the perspective transformation matrix and $J$ is the Jacobian of the projection matrix. The contribution of projected 2D Gaussian to each pixel $(x,y)$ is computed as: $\alpha_i = \exp(-\frac{1}{2}(p-\mu'_i)^T(\Sigma'_i)^{-1}(p-\mu'_i))$,
where $p$ is the pixel coordinate and $\mu'_i$ is the projected mean. The final color at each pixel is obtained by alpha compositing the contributions from all Gaussians, sorted by depth: $C = \sum_{i=1}^M T_i \alpha_i c_i$,
where $T_i = \prod_{j<i}(1 - \alpha_j)$ is the accumulated transmittance, $c_i$ is the view-dependent color computed from spherical harmonics coefficients, and $M$ is the number of Gaussians contributing to the pixel.

\subsection{Reconstruction Uncertainty Prediction}

Given the input images $\{I_n\}_{n=1}^N$, the goal is to optimize the unsupervised reconstruction uncertainty predictor \emph{RUP} through 3DGS reconstruction to identify transient distractors without explicit supervision as shown in Fig.~\ref{fig:pipeline}. Following prior research~\cite{martin2021nerf, ren2024nerf, kulhanek2024wildgaussians} we employ uncertainty modeling techniques, with significant modifications. \emph{RUP} is trained to identify transient objects without explicit supervision, purely from the reconstruction objectives. Several recent works~\cite{dahmani2024swag, sabour2024spotlesssplats, kulhanek2024wildgaussians, gu2025egolifter, xu2024wild} follow a similar approach, demonstrating the effectiveness of this optimization in handling dynamic scenes. As in WildGaussians~\cite{kulhanek2024wildgaussians} (and, in contrast to NeRF counterparts~\cite{martin2021nerf, ren2024nerf}) every iteration we update both Gaussian Splatting and \emph{RUP} weights. Additionally, we detach masks when updating Gaussian Splatting and detach reconstructed images when updating \emph{RUP}.

Following~\cite{kulhanek2024wildgaussians, sabour2024spotlesssplats}, we reformulate the transient detection problem as a semantic feature classification task. This approach leverages pre-trained foundation models to extract rich semantic features from images. By doing so, it enables our system to make decisions based on high-level semantic understanding, rather than relying solely on color information. This semantics-aware approach is more robust in distinguishing between static and transient objects than traditional color-based methods.

\subsubsection{Feature Extraction}
For each training iteration, we extract DINOv2 features~\cite{oquab2023dinov2} from both the input image $I$ and the corresponding rendering $\hat{I}$, producing feature maps $f, \hat{f}$ respectively. We choose DINOv2 for several key reasons: (1) its self-supervised training enables robust semantic understanding without class-specific biases, (2) it demonstrates strong performance in distinguishing object boundaries and semantic regions even for previously unseen objects, (3) compared to alternatives like DIFT~\cite{tang2023emergent} features, DINOv2 offers significantly faster computation times, making it more practical for iterative training processes. These features serve as a robust foundation, enabling \emph{RUP} to make accurate decisions about scene dynamics without explicit supervision.

\subsubsection{Transient 2D Uncertainty Modeling}
As previously discussed, most methods detect transient objects by utilizing reconstruction errors. For example, NeRF On-the-go~\cite{ren2024nerf} considers RGB residuals:
\begin{equation}
    R = ||\hat{I} - I||_2.
\end{equation}
It assumes that the residuals follow a normal distribution:
\begin{equation}
    p(R | \sigma) = \frac{1}{\sqrt{2 \pi \sigma^2}} \exp\left(-\frac{R^2}{2\sigma^2}\right).
\end{equation}
Therefore, we can obtain negative log likelihood:
\begin{equation}
    \mathcal{L}_u = \frac{R^2}{2\sigma^2} + \log \sigma + \frac{\log 2\pi}{2}.
\end{equation}

Although the approach is reasonable, RGB residuals lack robustness. In particular, high-frequency objects often result in high reconstruction errors, producing incorrect misclassification. Similarly, dynamic objects with colors similar to the background may be classified as static. While DSSIM or DINOv2 cosine distance can mitigate some errors, they introduce their own limitations. DSSIM residuals are susceptible to similar errors as RGB residuals, though to a lesser degree. The DINOv2 cosine distance, while highly robust, suffers from low resolution. Upsampling models, such as FeatUP~\cite{fu2024featup}, can address this issue, though they introduce upsampling artifacts.

This motivates a new multivariate formulation of uncertainty modeling. Let the residual be a 2-dimensional vector:
\begin{equation}
    R = \begin{bmatrix} R_1 \\ R_2\end{bmatrix},
\end{equation}
where $R_1$ and $R_2$ correspond to different similarity metrics: 1) DINOv2 cosine distance defined like in WildGaussians, except we upscale it with FeatUP~\cite{fu2024featup} and 2) DSSIM.
We consider a multivariate normal distribution with zero mean and covariance matrix $\Sigma$:
\begin{equation}
    p(R) = \frac{1}{(2\pi) \sqrt{|\Sigma|}} \exp\left(-\frac{1}{2} R^T \Sigma^{-1} R\right),
\end{equation}
where the covariance matrix is given by
\begin{equation}
    \Sigma = \begin{bmatrix} \sigma_1^2 & \rho \sigma_1 \sigma_2 \\ \rho \sigma_1 \sigma_2 & \sigma_2^2 \end{bmatrix}.
\end{equation}
The negative log-likelihood function becomes:
\begin{equation}
    \mathcal{L}_u = -\log p(R) = \frac{1}{2} R^T \Sigma^{-1} R + \frac{1}{2} \log |\Sigma| + \log 2\pi.
    \label{likelihood}
\end{equation}

In contrast to previous works ~\cite{martin2021nerf, ren2024nerf, kulhanek2024wildgaussians} we predict three parameters --- $\sigma_1$, $\sigma_2$, $\rho$ instead of a single $\sigma$. This allows us to combine information about both residuals.

There is a problem in this derivation due to the assumption that residuals to be strictly positive. This implies that our distribution represents only the positive quadrant of the bivariate normal distribution. While this is relevant for the subsequent derivations, it does not affect the likelihood, as it only introduces a scaling factor, which can be ignored during optimization. Interestingly, this is not important for the one-dimensional case, where likelihood depends only on the absolute value of the residual.

We train a neural network that takes DINOv2 features from a reference image as input and makes a per pixel prediction of $\Sigma$, and use our likelihood term in Eq.~\ref{likelihood} as a loss function.  It should be noted that $\Sigma$ can be noninvertible when $\sigma_i=0$ or $\rho=\pm 1$. Although we predict $\sigma_i$ using a softplus nonlinearity, and $\rho$ using a tanh nonlinearity to avoid undesirable values, in practice this can lead to 1) numerical instabilities and 2) undesirable values due to the discrete representation of numerical values. The second problem is easy to solve with clamping, in our experience, the first problem was mostly solved by introducing normalization layers into the architecture of the neural network.

\subsubsection{Model Architecture}
Given that our training objective is considerably more challenging than the one-dimensional modeling of WildGaussians~\cite{kulhanek2024wildgaussians}, our model requires a larger architecture. However, this also offers an advantage over previous methods, as we can use simple upscale layers to make our prediction denser without sacrificing local/nonlocal smoothing. The details of architecture are provided in the Supplementary Material.

\subsubsection{Binary Mask}

One approach to obtaining a binary mask using the modeled uncertainty is to set a threshold on one of the predicted values or define a new criterion:

\begin{equation}
    M = \mathbb{I}(f(\sigma_1, \sigma_2, \rho) > C),
\end{equation}
where $\mathbb{I}$ is the indicator function, $f(\sigma_1, \sigma_2, \rho)$ is a chosen criterion and $C$ is a threshold chosen as a hyperparameter.

However, this methodology has notable limitations when applied to the reconstruction of geometrically complex static structures. In such cases, even static objects may produce substantial residuals, leading to their misclassification as dynamic elements and their subsequent masking. This misclassification ultimately degrades the reconstruction quality of static scene components. To address these limitations, we introduce necessary regularization constraints.

We note that, even though we train our \emph{RUP} only on a reference images, we can also obtain a per pixel uncertainty prediction $\hat{\Sigma}$ using an image reconstructed by Gaussian Splatting model. Because our model relies on semantic information of DINOv2 features, we should expect it to make a similar prediction in the static areas and a different one in areas corresponding to the dynamic objects. To estimate this discrepancy, we calculate the Kullback-Leibler (KL) divergence $D_{KL}(\mathcal{N}(0, \Sigma) || \mathcal{N}(0, \hat{\Sigma}))$, which takes following form for two normal distributions:

\begin{equation}
    D_{KL}(\mathcal{N}(0, \Sigma)|| \mathcal{N}(0, \hat{\Sigma})) = \frac{1}{2}(tr(\hat{\Sigma}^{-1}\Sigma) - ln(\frac{|\Sigma|}{|\hat{\Sigma}|}) - 2).
    \label{KL_criterion}
\end{equation}

\begin{figure}[t]
    \centering
    \includegraphics[width=1.\linewidth]{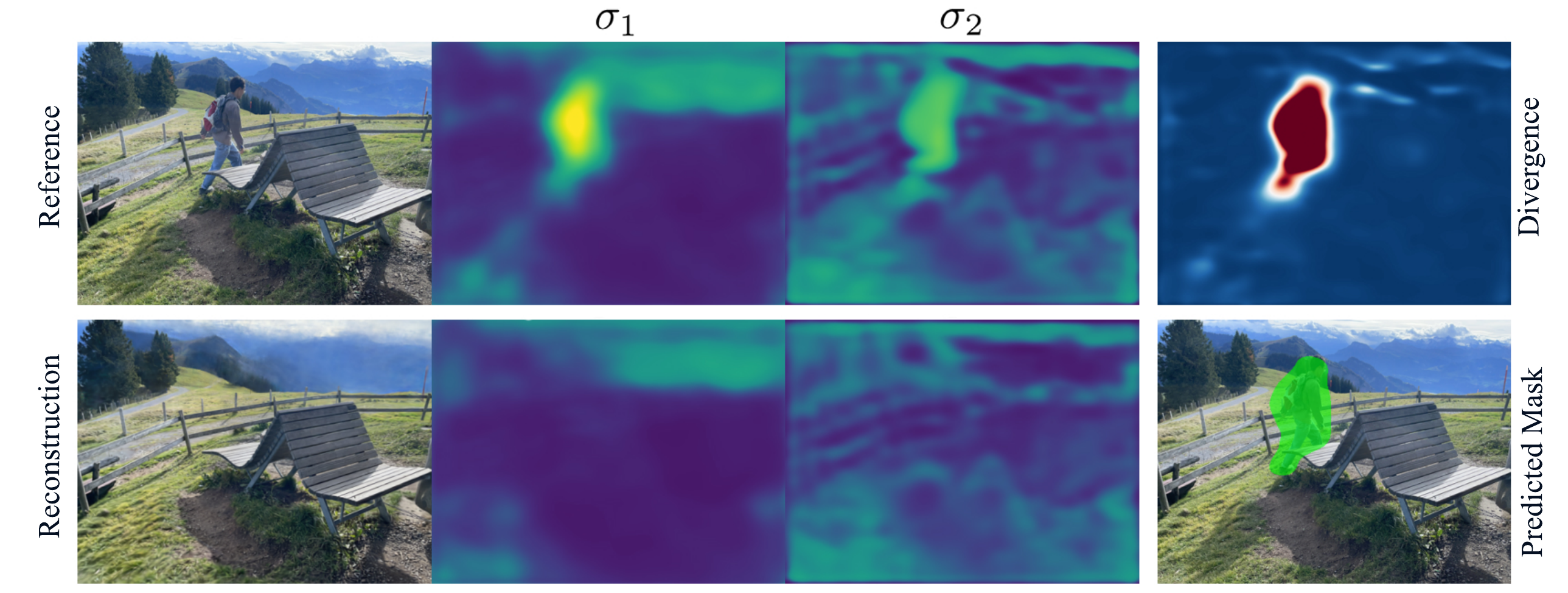}
    \caption{During the initial stages of reconstruction, \emph{RUP} predicts high uncertainty in challenging regions such as backgrounds or high frequency details. However, since \emph{RUP} relies exclusively on semantic information, calculating the divergence between reference uncertainty $\Sigma$ and reconstructed uncertainty $\hat{\Sigma}$ effectively suppresses these artifacts. Areas with divergence values above the threshold are highlighted in red, while the final predicted transient mask by \emph{RUP} is shown in green.
    }
    \vspace{-0.5cm}
    \label{fig:consistency}
\end{figure}

Fig.~\ref{fig:consistency} illustrates how this approach reduces false classifications in static regions. Unlike previous methods that obtain masks by estimating uncertainty, we instead utilize divergence. This allows us to incorporate additional information to enhance the consistency of our masks. Hence, our binary masks are obtained based on the new criterion:
\begin{equation}
    M = \mathbb{I}(D_{KL} > C).
    \label{divergence_criterion}
\end{equation}

Notably, formula \eqref{KL_criterion} is not exact since we are working with a folded multivariate distribution. We can add an ad hoc assumption that we observe only the absolute value of the residuals and its sign is random. Nevertheless, it has proven to be a useful heuristic. We leave a more careful derivation for future work.

\subsubsection{Training Stability}

During 3DGS optimization, there are periods when renders may be unreliable, particularly at the beginning of training and after each opacity reset. Building on ~\cite{kulhanek2024wildgaussians}, we address this by implementing two key strategies. First, we delay the start of \emph{RUP} training until the 3DGS optimization has completed its first 500 iterations, ensuring the initial scene reconstruction has reached sufficient quality. Second, after each opacity reset, we temporarily pause the \emph{RUP} optimization for 250 iterations while keep training 3DGS, allowing the reconstruction to stabilize before resuming transient detection. We also use scheduled sampling technique from SpotLessSplats~\cite{sabour2024spotlesssplats}.

\subsubsection{Mask Dilation}

We also dilate our masks, depending on the resolution of the scene. This dilation step serves multiple purposes, primarily covering shadows and reflections caused by objects. These modifications ensure robust training and accurate detection of transient objects in diverse dynamic scenes.

\subsection{Transient Mask Refinement}

Our transient area detection pipeline is robust for current benchmarks, where transient objects are always dynamic and change their positions from frame to frame. However, for semi-transient objects, which may not change positions for some frames, it fails and masks only parts of the video when they are dynamic. To address this issue, we introduce a mask propagation process that refines transient masks into temporally consistent, accurate masks with high-resolution boundaries across the entire video sequence through refinement and propagation. Each segmentation consists of a binary mask that defines the object's spatial extent and a unique label, consistent throughout the video sequence.

\subsubsection{Spatial Refinement}
\label{sec:spatial}
We use the Segment Anything Model (SAM) \cite{kirillov2023segment} to refine our transient maps, $P_i$, into more precise masks, $M'_i$. For each connected component $C_i^k$ in $P_i$, we sample up to ten points as prompts for SAM, leveraging its ability to generate high-quality segmentations from sparse inputs to extract a set of object-aware masks $M'_i = \{M'{_i^j}\}_{j=1}^{L_i}$, where $L_i$ is the number of predicted masks for image $I_i$. Due to potential inaccuracies in the boundaries of our masks, some sampled points might occasionally fall on the background rather than the object itself (\eg, a point sampled between the legs of a person). To address this, we filter the predicted masks based on their local coverage score:
\begin{equation}
    CS_{\text{local},i} = \frac{|P_i \cap M'{_i^j}|}{|M'{_i^j}|}. \label{eq:cs_local}
\end{equation}
We keep only masks that satisfy $CS_{\text{local},i} > \lambda_{\text{cov}}^{\text{ref}}$, forming the refined set $M'_i = \{M'{_i^j} | CS_{\text{local},i} > \lambda_{\text{cov}}^{\text{ref}}\}$.

\subsubsection{Temporal Refinement}
\label{sec:temporal}
To address potential false negatives, we propagate the refined masks, $\{M'_i\}_{i=1}^N$, throughout the video using SAM2~\cite{ravi2024sam} to obtain more consistent masks, $\{M_i\}_{i=1}^N$. Our propagation process consists of three stages:

\begin{enumerate}
\item \underline{Forward Propagation}: Iterating from the first frame to the last, propagating the segmentation masks forward.
\item \underline{Backward Propagation}: Iterating from the last frame to the first, propagating information from future frames backward.
\item \underline{Final Propagation}: A final first-to-last pass, considering both past and future frames as context, which helps to resolve temporal inconsistencies.
\end{enumerate}

To manage computational resources efficiently, we introduce a memory size parameter, $N_m$, which limits the number of frames considered during propagation. At each step, we maintain and use segmentations from  $N_m$ nearest frames, balancing temporal consistency with memory constraints.

During propagation, we manage mask intersections to ensure consistent segmentation. For any pair of masks $M_i^{l}$ and $M_i^{m}$ where $IoU(M_i^{l}, M_i^{m}) > \lambda_{merge}$, we merge them into a single mask, assigning the lower of the two original labels to maintain consistency.

\subsubsection{Dynamic Object Filtration}
\label{sec:filtration}
To filter out false positive transients and ensure robust detection, we introduce the Stability Ratio (SR) metric, which combines spatial overlap accuracy and temporal consistency. For each detected object, the SR is calculated as $SR = \frac{1}{N} \sum_{i=1}^{N} (R_i \cdot CS_{\text{global},i})$, where $N$ is the number of valid frames, $R_i$ is the mean value of the absolute difference between ground truth and rendered images within the masked region in frame $i$, and $CS_{\text{global},i} = |P_i \cap M_i|/|M_{\text{max}}|$ is the global coverage score. Here, $P_i$ represents the prompt mask in frame $i$, $M_i$ is the segmentation mask, and $M_{\text{max}}$ is the maximum size of the object mask across all frames. This global score evaluates the object's consistency relative to its largest observed size.
A frame is considered valid and contributes to the SR calculation only if its local coverage score (Eq.~\ref{eq:cs_local}) exceeds the validation threshold $\lambda_{\text{cov}}^{\text{val}}$. Objects with $SR$ below a threshold $\lambda_{SR}$ are filtered out as potential false detections. This dual coverage score system ensures that objects maintain both spatial accuracy through local coverage and temporal consistency through global coverage and difference image values.

\subsection{Artifact-Free Reconstruction}
3DGS tends to generate floating artifacts ("floaters") near the camera, particularly in challenging regions like those identified by transient masks. These artifacts can saturate gradients, thereby degrading overall reconstruction quality. We address this issue through depth-aware regularization.

We render the depth $D$ for each pixel using alpha compositing, similar to color rendering: $D = \sum_{i=1}^M T_i \alpha_i d_i$,
where $d_i$ is the depth value of the i-th Gaussian, $T_i$ is the accumulated transmittance, and $\alpha_i$ is the opacity value. To suppress floating artifacts while preserving sharp depth discontinuities at object boundaries, we apply anisotropic total variation (TV) regularization to the rendered depth map: $\mathcal{L}_{\text{depth}} = \text{mean}(|\nabla_x D|) + \text{mean}(|\nabla_y D|)$,
where $\nabla_x$ and $\nabla_y$ are spatial gradients in $x$ and $y$ directions respectively.
\subsection{Masked Gaussian Splatting Optimization}

The final step involves training the Gaussian Splatting model with the obtained masks $\{M_i\}_{i=1}^N$ for transients. Let $M_i$ be the binary mask for frame $i$, defined as:

\begin{equation}
M_i(x, y) = \begin{cases}
1 & \text{if $(x,y)$ is in an occluded area}, \\
0 & \text{if $(x,y)$ is in a static area},
\end{cases}
\end{equation}
where $(x,y)$ represents pixel coordinates in the image.
We apply binary dilation to $M_i$ for $N_e$ iterations, yielding $M^*_i$. This operation creates a buffer zone around detected dynamic objects, improving the robustness of our static scene reconstruction. The final loss for 3DGS is:
\begin{equation}
\begin{split}
    \mathcal{L}_{\text{masked}} = \lambda_{\text{SSIM}} \cdot L_{\text{SSIM}}(I_i \odot \overline{M}^*_i, \hat{I_i} \odot \overline{M}^*_i) + \\
    \lambda_{\text{L1}} \cdot \left|\left| \overline{M}^*_i \odot (I_i - \hat{I_i}) \right| \right|_1 + \lambda_{\text{depth}} \mathcal{L}_{\text{depth}},
    \label{masked_loss}
\end{split}
\end{equation}
where $I_i$, $\hat{I_i}$ are reference images and their reconstructions, $\odot$ is the Hadamard product, $\| \cdot \|_1$ is L1 norm, $L_{SSIM}$ is a structural similarity loss, $\overline{M}^*_i$ is a negation of $M^*_i$ that represents a static background and $\lambda_{\text{SSIM}}$, $\lambda_{\text{L1}}$ and  $\lambda_{\text{depth}}$ are weighting factors.

This formulation allows the model to focus on static scene elements, effectively handling dynamic objects in the reconstruction process. By integrating these steps, our method reconstructs static scenes robustly from unconstrained videos while effectively handling transient distractors.

\section{Experiments}

\begin{figure}
  \centering
  \includegraphics[width=\linewidth]{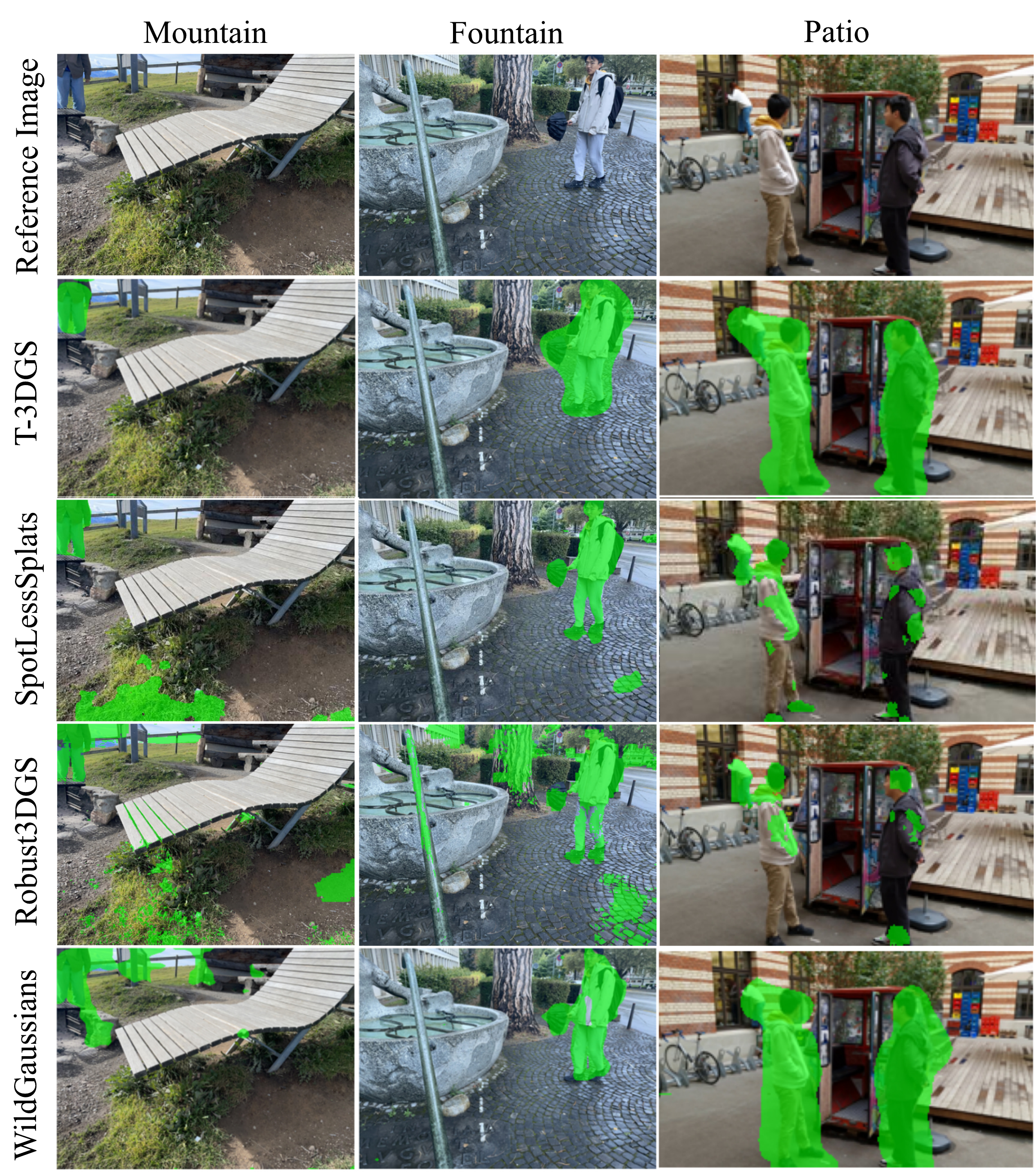}
  \vspace{-0.2cm}
  \caption{Qualitative results on the \emph{On-the-go} dataset. Our method outperforms existing approaches in detecting transient objects. Predicted transient masks are shown in green.}
  \label{fig:onthego_vis}
\end{figure}

\begin{figure*}[hbt!]
  \centering
  \includegraphics[trim={0 9.3cm 0 0},clip,width=\linewidth]{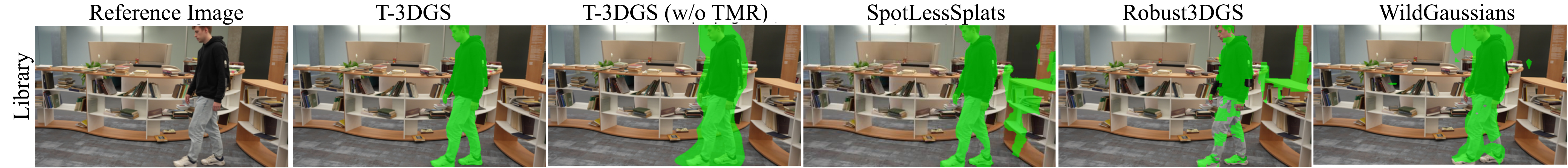}
  \caption{Qualitative results on the \emph{T-3DGS} dataset. Our method produces cleaner transient masks and further refines them using the (\emph{TMR}) module.}
  \vspace{-0.3cm}
  \label{fig:t3dgs_vis_cut}
\end{figure*}

\begin{table*}[hbt!]
    \centering
    \scalebox{.474}{
    \begin{tabular}{l|ccc|ccc|ccc|ccc|ccc|ccc|ccc}
        & \multicolumn{3}{|c|}{Mountain} & \multicolumn{3}{c|}{Fountain} & \multicolumn{3}{c|}{Corner} & \multicolumn{3}{c|}{Patio} & \multicolumn{3}{c|}{Spot} & \multicolumn{3}{c|}{Patio High} & \multicolumn{3}{c}{Mean} \\
        & PSNR $\uparrow$ & SSIM $\uparrow$ & LPIPS $\downarrow$ & PSNR $\uparrow$ & SSIM $\uparrow$ & LPIPS $\downarrow$ & PSNR $\uparrow$ & SSIM $\uparrow$ & LPIPS $\downarrow$ & PSNR $\uparrow$ & SSIM $\uparrow$ & LPIPS $\downarrow$ & PSNR $\uparrow$ & SSIM $\uparrow$ & LPIPS $\downarrow$& PSNR $\uparrow$ & SSIM $\uparrow$ & LPIPS $\downarrow$ & PSNR $\uparrow$ & SSIM $\uparrow$ & LPIPS $\downarrow$ \\
        \hline
        NeRF On-the-go~\cite{ren2024nerf} & 20.15 & 0.64 & 0.26 & 20.11 & 0.61 & 0.31 & 24.22 & 0.81 & 0.19 & 20.78 & 0.75 & 0.22 & 23.33 & 0.79 & 0.19 & 21.41 & 0.72 & 0.24 & 21.67 & 0.72 & 0.24 \\
        \hline
        3DGS~\cite{kerbl20233d} & 19.40 & \underline{0.66} & \textbf{0.21} & 19.96 & \textbf{0.66} & \textbf{0.19} & 20.90 & 0.71 & 0.24 & 17.48 & 0.70 & 0.20 & 20.77 & 0.69 & 0.32 & 17.29 & 0.60 & 0.36 & 19.30 & 0.67 & 0.25 \\
        Robust3DGS~\cite{ungermann2024robust} & 16.97 & 0.61 & 0.31 & 18.18 & 0.59 & 0.32 & 23.47 & 0.85 & \underline{0.10} & 21.33 & \underline{0.85} & \textbf{0.07} & 22.61 & \underline{0.88} & \underline{0.12} & 21.81 & \underline{0.82} & \textbf{0.17} & 20.73 & 0.77 & 0.19 \\
        WildGaussians~\cite{kulhanek2024wildgaussians} & 20.77 & \underline{0.70} & 0.23 & \underline{20.74} & \underline{0.67} & \underline{0.21} & \underline{25.79} & \underline{0.88} & \textbf{0.09} & \textbf{21.77} & \underline{0.85} & \textbf{0.07} & \underline{24.39} & \underline{0.88} & \textbf{0.10} & \underline{22.36} & 0.80 & \textbf{0.17} & \underline{22.64} & \underline{0.80} & \textbf{0.15} \\
        SpotLessSplats~\cite{sabour2024spotlesssplats} & \textbf{21.25} & 0.66 & 0.24 & 20.49 & 0.63 & 0.24 & 25.59 & 0.85 & 0.12 & 21.13 & 0.80 & \underline{0.08} & 24.13 & 0.78 & 0.18 & 22.18 & 0.76 & \underline{0.20} & 22.46 & 0.75 & 0.18 \\
        \hline
        \textbf{Ours} & \underline{21.11} & \textbf{0.71} & \underline{0.22} & \textbf{20.94} & \textbf{0.69} & \underline{0.21}  & \textbf{26.46} & \textbf{0.90} & 0.12 & \underline{21.95} & \textbf{0.87} & 0.10 & \textbf{25.78} & \textbf{0.90} & \underline{0.12} & \textbf{22.76} & \textbf{0.83} & \textbf{0.17} & \textbf{23.17} & \textbf{0.82} & \underline{0.16} \\
    \end{tabular}
    }
    \caption{Quantitative comparison on the \emph{On-the-go} dataset~\cite{ren2024nerf}.}
    \vspace{-0.2cm}
    \label{tab:on-the-go}
\end{table*}

\begin{table*}[hbt!]
    \centering
    \scalebox{.55}{
    \begin{tabular}{l|ccc|ccc|ccc|ccc|ccc|ccc}
        & \multicolumn{3}{|c|}{Lab 1} & \multicolumn{3}{c|}{Lab 2} & \multicolumn{3}{c|}{Library} & \multicolumn{3}{c|}{Anti-Stress} & \multicolumn{3}{c|}{Office} & \multicolumn{3}{c}{Mean} \\
        & PSNR $\uparrow$ & SSIM $\uparrow$ & LPIPS $\downarrow$ & PSNR $\uparrow$ & SSIM $\uparrow$ & LPIPS $\downarrow$ & PSNR $\uparrow$ & SSIM $\uparrow$ & LPIPS $\downarrow$ & PSNR $\uparrow$ & SSIM $\uparrow$ & LPIPS $\downarrow$ & PSNR $\uparrow$ & SSIM $\uparrow$ & LPIPS $\downarrow$ & PSNR $\uparrow$ & SSIM $\uparrow$ & LPIPS $\downarrow$ \\
        \hline
        3DGS~\cite{kerbl20233d} & 24.49	& 0.91 & - & 20.42 & 0.87 & - & 20.08 & 0.89 & - & 20.45 & 0.86 & - & 26.96 & 0.94 & - & 22.48 & 0.89 & - \\
        Robust3DGS~\cite{ungermann2024robust} & 25.35 & \underline{0.93} & 0.09 & \underline{24.74} & \underline{0.92} & 0.10 & 24.33 & \underline{0.93} & \underline{0.08} & 22.95 & 0.91 & 0.10 & 28.52 & \textbf{0.96} & 0.06 & 25.18 & \underline{0.93} & 0.09 \\
        WildGaussians~\cite{kulhanek2024wildgaussians} & \underline{25.71} & 0.92 & \underline{0.08} & 23.68 & 0.91 & 0.11 & \underline{24.65} & 0.92 & 0.09 & 21.69 & 0.89 & \underline{0.09} & 28.89 & \underline{0.95} & \underline{0.04} & 24.92 & 0.92 & \underline{0.08} \\
        SpotLessSplats~\cite{sabour2024spotlesssplats} & 25.28 & 0.91 & \underline{0.08} & 24.63 & 0.90 & \underline{0.09} & 24.11 & 0.91 & 0.09 & 22.22 & 0.90 & 0.10 & 28.08 & 0.92 & 0.05 & 24.86 & 0.91 & \underline{0.08} \\
        \hline
        \textbf{Ours w/o TMR} & 25.77 & \underline{0.93} & 0.09 & \underline{24.67} & \underline{0.92} & 0.10 & 24.67 & 0.93 & \underline{0.08} & \underline{24.07} & \underline{0.92} & \underline{0.09} & \underline{29.36} & \underline{0.95} & 0.05 & \underline{25.71} & \underline{0.93} & \underline{0.08} \\
        \textbf{Ours w/ TMR} & \textbf{27.76} & \textbf{0.96} & \textbf{0.02} & \textbf{25.54} & \textbf{0.93} & \textbf{0.06} & \textbf{28.25} & \textbf{0.97} & \textbf{0.02} & \textbf{29.01} & \textbf{0.96} & \textbf{0.02} & \textbf{29.85} & \textbf{0.96} & \textbf{0.02} & \textbf{28.08} & \textbf{0.96} & \textbf{0.03}
    \end{tabular}}
    \caption{Quantitative comparison on the \emph{T-3DGS} dataset.}
    \vspace{-0.3cm}
    \label{tab:t3dgs}
\end{table*}

We evaluate our proposed T-3DGS model on various datasets captured in uncontrolled settings and filled with diverse distractors. We perform qualitative and quantitative comparisons against state-of-the-art methods. Finally, we provide an ablation study of architectural and loss function choices. We discuss the limitations of the proposed method in the Supplementary Material.

\noindent\textbf{Datasets.} We evaluate our model on three challenging datasets. The \emph{NeRF On-the-go dataset}~\cite{ren2024nerf} contains four outdoor and two indoor sparsely captured scenes with different levels of occlusion (from 5\% to over 30\%) and minimal appearance changes. The \emph{RobustNeRF dataset}~\cite{sabour2023robustnerf} contains five indoor scenes with unintentional changes during the capture process. These changes include transient objects that appear and disappear without a consistent temporal order, as well as dynamic objects (e.g., floating balloons). Additionally, we introduce our novel \emph{T-3DGS dataset}. The dataset contains 5 densely captured indoor scenes. Generally, dynamic objects in our videos are walking people and various small objects. However, unlike previous datasets, all scenes incorporate challenging cases, including transient, semi-transient, and slow-moving objects.

\noindent\textbf{Baselines.}
We compare our model against vanilla 3D Gaussian Splatting~\cite{kerbl20233d} and the current state-of-the-art method, SpotLessSplats~\cite{sabour2024spotlesssplats}. We further include WildGaussians~\cite{kulhanek2024wildgaussians} and Robust3DGS~\cite{ungermann2024robust} as baselines. To compare different models, we use commonly used PSNR, SSIM~\cite{wang2004image} and LPIPS metrics for evaluation.

\noindent\textbf{Implementation details.}
All our experiments are conducted in accordance with the training setup from the official 3DGS implementation. We train our models for 30K iterations, using the Adam optimizer with a learning rate of 1e-3 for the \emph{RUP}. The depth regularization loss $\mathcal{L}_{\text{depth}}$ is activated after the first 500 iterations, allowing the 3DGS to establish initial geometry reconstruction. For the experiments with mask propagation, we first train the \emph{RUP} for 7000 iterations. At that point, we pause the training to propagate the transient masks. Subsequently, we initiate a new training procedure using the propagated masks, keeping all other parameters the same as the original training setup. We dilate all our masks by 10 pixels, except for the Patio scene, where we use the original mask due to its low resolution.

\subsection{Quantitative Comparisons}

We evaluate our model on all three datasets. We report results on \emph{On-the-go} and \emph{T-3DGS} datasets in Tab.~\ref{tab:on-the-go} and \ref{tab:t3dgs}, respectively, and we move the evaluation results of \emph{RobustNeRF} dataset to the Supplementary Material as it presents the least challenge. As shown in Tab.~\ref{tab:on-the-go} and \ref{tab:t3dgs}, our method generally outperforms current SOTA methods. In particular, our method is robust to changes in distant and high-frequency details. In Tab.~\ref{tab:on-the-go} we run our method directly on masks predicted by \emph{RUP} module without mask propagator.

While current SOTA methods struggle to detect semi-transient objects (Tab.~\ref{tab:t3dgs}), our proposed transient network \emph{RUP} achieves higher performance by minimizing false predictions. The integration of the SAM-based mask propagation \emph{TMR} module further enhances our results in scenes containing semi-transient objects, providing more accurate and reliable reconstructions.

\subsection{Qualitative Comparisons}

For qualitative evaluation, we compare our method to SpotLessSplats~\cite{sabour2024spotlesssplats}, Robust3DGS~\cite{ungermann2024robust}, and WildGaussians~\cite{kulhanek2024wildgaussians}. Fig.~\ref{fig:onthego_vis} and~\ref{fig:t3dgs_vis_cut} demonstrate that our method minimizes false negatives and effectively detects transients. For example, in the On-the-go dataset, most methods struggle with high-frequency details and distant objects, as these elements are typically reconstructed more slowly than the rest of the scene, leading to inaccuracies in RGB residual-based approaches. However, due to our robust loss function, such artifacts are largely eliminated from our dynamic maps. Notably, SpotLessSplats uses features obtained from higher-resolution images, while we extract features at a lower resolution, the same resolution used for training 3DGS.

For our \emph{T-3DGS} dataset, we additionally utilize the SAM-based mask propagation module to propagate object-aware masks for semi-transient objects, as shown in Fig.~\ref{fig:t3dgs_vis_cut}. Although most methods would theoretically benefit from this technique, our masks are of higher quality and result in fewer incorrect detections. Applying mask propagation to other methods may introduce error propagation, as demonstrated in the Supplementary Material.

\begin{table}[t]
    \centering
    \scalebox{0.7}{
    \begin{tabular}{l|cc}
        & \multicolumn{2}{|c}{\emph{On-the-go dataset}} \\
        & PSNR $\uparrow$ & SSIM $\uparrow$ \\
        \hline
        GT masks w/o $\mathcal{L}_{\text{depth}}$ & 22.84 & 0.82 \\
        GT masks w/ $\mathcal{L}_{\text{depth}}$ and dilation & 23.43 & 0.81 \\
        \hline
        Ours w/o dilation and $\mathcal{L}_{\text{depth}}$ & 22.60 & 0.80 \\
        Ours w/o dilation & 22.88 & 0.80 \\
        \textbf{Ours (full)} & 23.41 & 0.81
    \end{tabular}}
    \vspace{-0.2cm}
    \caption{We evaluate the importance of each component of our method on the \emph{On-the-go} dataset. We report the average performance across all scenes.}
    \vspace{-0.3cm}
    \label{tab:ablation}
\end{table}

\subsection{Ablation Study}

We present ablation results in Table~\ref{tab:ablation} for the \emph{On-the-go} dataset, excluding the Patio scene due to its low resolution. We evaluate our method under the following conditions: (1) without mask dilation, (2) without mask dilation and $\mathcal{L}_{\text{depth}}$, and (3) with both components enabled. Additionally, we report results obtained with ground truth masks while separately disabling $\mathcal{L}_{\text{depth}}$ and mask dilation by 10 pixels. Even when using ground truth masks, dilation noticeably enhances performance. This contradicts the assumptions made by NeRF-HuGS~\cite{chen2024nerf} and Robust3DGS~\cite{ungermann2024robust}, as exact masks do not yield optimal performance metrics. Furthermore, mask dilation aids \emph{RUP} training by ensuring that all transient objects are fully covered. We also note that our results align very closely with those obtained using GT masks, suggesting that more challenging datasets are required.

\section{Conclusion}

In this work, we have presented the novel \textit{T-3DGS} method for 3D scene reconstruction using Gaussian Splatting by effectively filtering out foreground dynamic distractors from input videos. By integrating an unsupervised classification network with bivariate uncertainty modeling, KL divergence regularization, and a mask propagation strategy, our method achieves superior temporal coherence and boundary accuracy. Evaluations on both sparsely and densely captured datasets confirm significant improvements over state-of-the-art approaches. We believe our method represents a significant step toward the broader adoption of 3DGS for robust 3D scene reconstruction from real-world videos captured in uncontrolled settings.

\clearpage
{
    \small
    \bibliographystyle{ieeenat_fullname}
    \bibliography{main}
}

\clearpage
\appendix
\clearpage
\setcounter{section}{0}
\maketitlesupplementary
\renewcommand*{\thesection}{\Alph{section}}

\section{Limitations}
We use features upscaled by FeatUP~\cite{fu2024featup} to compute cosine distance, and while it is better than simple bilinear interpolation, it is relatively slow and gives fairly noisy results. Utilizing alternative ways to measure per pixel errors might improve both speed and accuracy of the method.
Additionally, the temporal refinement process is constrained by a memory window of $N_m$ frames, which means that if an object disappears for more than $N_m$ frames and then reappears, it will be treated as a new instance with a different label. This can lead to inconsistent tracking and potentially affect the filtering process, especially for semi-transient objects that may temporarily leave the scene. Furthermore, our current filtering approach using global coverage scores may incorrectly filter out valid dynamic objects that undergo significant size changes, such as objects moving towards or away from the camera, or those experiencing perspective changes. We leave it as a future work.

\section{Additional Implementation Details}
In Sec.~3.3.1, for mask filtering and refinement, we set $\lambda_{\text{cov}}^{\text{ref}} = 0.7$ for initial mask refinement and $\lambda_{\text{cov}}^{\text{val}} = 0.7$ for validation during object filtration. For temporal refinement in Sec.~3.3.2, we set the memory size parameter $N_m = 10$, which controls the number of frames considered during mask propagation. For the final mask dilation step, we perform $N_e = 5$ iterations of binary dilation. In addition, the mask merging threshold $\lambda_{merge}$ is set to 0.9, and the stability ratio threshold $\lambda_{SR}$ to 0.08 in Sec.~3.3.3.

Our model consists of repeating blocks. We fist use bilinear interpolation to increase the resolution of our features by two. We then apply a simple 3 by 3 convolutional layer that also decreases feature size by a factor of two. We then apply layer normalization followed by the GELU non-linearity. We repeat this sequence three times. After that we project our features with 1 by 1 convolution to obtain logits. We use softplus for $\sigma_1$, $\sigma_2$ and tanh for $\rho$. The normalization layer is crucial for improving the numerical stability that arises due to matrix $\Sigma$ being potentially ill-conditioned.

\section{Evaluation on RobustNeRF Dataset}
We evaluate our method on the \emph{RobustNeRF} dataset~\cite{sabour2023robustnerf}. As shown in Tab.~\ref{tab:robustnerf} our method generally outperforms 3DGS~\cite{kerbl20233d}, Robust 3DGS~\cite{ungermann2024robust}, WildGaussians~\cite{kulhanek2024wildgaussians}, and shows similar performance compared to SpotLessSplats~\cite{sabour2024spotlesssplats}. We run our method directly on masks predicted by \emph{RUP} module without mask propagator (\emph{TMR}). Overall, the dataset does not appear to be sufficiently challenging to differentiate between the methods.

\begin{table*}[t]
    \centering
    \scalebox{.7}{
    \begin{tabular}{l|cc|cc|cc|cc|cc|cc}
        & \multicolumn{2}{|c|}{Android} & \multicolumn{2}{c|}{Statue} & \multicolumn{2}{c|}{Crab (1)} & \multicolumn{2}{c|}{Crab (2)} & \multicolumn{2}{c|}{Yoda} & \multicolumn{2}{c}{Mean} \\
        & PSNR $\uparrow$ & SSIM $\uparrow$ & PSNR $\uparrow$ & SSIM $\uparrow$ & PSNR $\uparrow$ & SSIM $\uparrow$ & PSNR $\uparrow$ & SSIM $\uparrow$ & PSNR $\uparrow$ & SSIM $\uparrow$ & PSNR $\uparrow$ & SSIM $\uparrow$ \\
        \hline
        NeRF On-the-go~\cite{ren2024nerf} & 23.50 & 0.75 & 21.58 & 0.77 & - & - & - & - & 29.96 & 0.83 & - & - \\
        \hline
        3DGS~\cite{kerbl20233d} & 23.51 & 0.81 & 21.35 & 0.84 & 30.39 & 0.94 & 31.53 & 0.92 & 29.80 & 0.92 & 27.32 & \underline{0.89} \\
        Robust 3DGS~\cite{ungermann2024robust} & 24.40 & \underline{0.83} & 22.10 & \underline{0.85} & 34.41 & \textbf{0.96} & 32.99 & \underline{0.93} & \underline{32.62} & \underline{0.93} & 29.30 & \textbf{0.90} \\
        WildGaussians~\cite{kulhanek2024wildgaussians} & \underline{24.89} & \underline{0.83} & \underline{22.69} & \textbf{0.87} & 30.16 & 0.93 & 31.11 & 0.91 & 30.50 & 0.91 & 27.87 & \underline{0.89} \\
        SpotLessSplats~\cite{sabour2024spotlesssplats} & 24.45 & 0.79 & 22.50 & 0.80 & \underline{35.45} & \underline{0.95} & \underline{33.29} & \textbf{0.94} & \textbf{33.55} & \textbf{0.94} & \textbf{29.85} & 0.88 \\
        \hline
        \textbf{Ours} & \textbf{25.10} & \textbf{0.84} & \textbf{22.90} & \textbf{0.87} & \textbf{34.25} & \underline{0.95} & \textbf{33.85} & \underline{0.93} & 32.45 & 0.93 & \underline{29.71} & \textbf{0.90} \\
    \end{tabular}
    }
    \caption{Quantitative comparison on the \emph{RobustNeRF} dataset~\cite{sabour2023robustnerf}.}
    \label{tab:robustnerf}
\end{table*}

\begin{table*}[t]
    \centering
    \scalebox{.75}{
    \vspace{-5pt}
    \begin{tabular}{l|cc|cc|cc|cc|cc|cc}
        & \multicolumn{2}{|c|}{Anti-Stress} & \multicolumn{2}{c|}{Lab (1)} & \multicolumn{2}{c|}{Lab (2) } & \multicolumn{2}{c|}{Library} & \multicolumn{2}{c|}{Office} & \multicolumn{2}{c}{Mean} \\
        & PSNR $\uparrow$ & SSIM $\uparrow$ & PSNR $\uparrow$ & SSIM $\uparrow$ & PSNR $\uparrow$ & SSIM $\uparrow$ & PSNR $\uparrow$ & SSIM $\uparrow$ & PSNR $\uparrow$ & SSIM $\uparrow$ & PSNR $\uparrow$ & SSIM $\uparrow$ \\
        \hline
        WildGaussians w/o TMR & 21.69 & 0.89 & 25.71 & 0.92 & 23.68 & 0.91 & 24.65 & 0.92 &  28.89 & 0.95 & 24.92 & 0.92 \\
        WildGaussians w/ TMR & 24.07 & 0.92 & 24.65 & 0.92 & 24.84 & \textbf{0.93} & 28.32 & \textbf{0.97} & 29.75 & \textbf{0.96} & 26.33 & 0.94 \\
        \hline
        \textbf{Ours w/ TMR} & \textbf{28.79} & \textbf{0.97} & \textbf{27.71} & \textbf{0.95} & \textbf{25.42} & \textbf{0.93} & \textbf{28.34} & \textbf{0.97} & \textbf{29.87} & \textbf{0.96} & \textbf{28.03} & \textbf{0.96} \\
    \end{tabular}
    }
    \caption{Evaluation of WildGaussians with \emph{TMR} module on the \emph{Transient-3DGS} dataset.}
    \label{tab:mask_comparison}
    \vspace{-5pt}
\end{table*}

\begin{figure*}
  \centering
  \includegraphics[width=\linewidth]{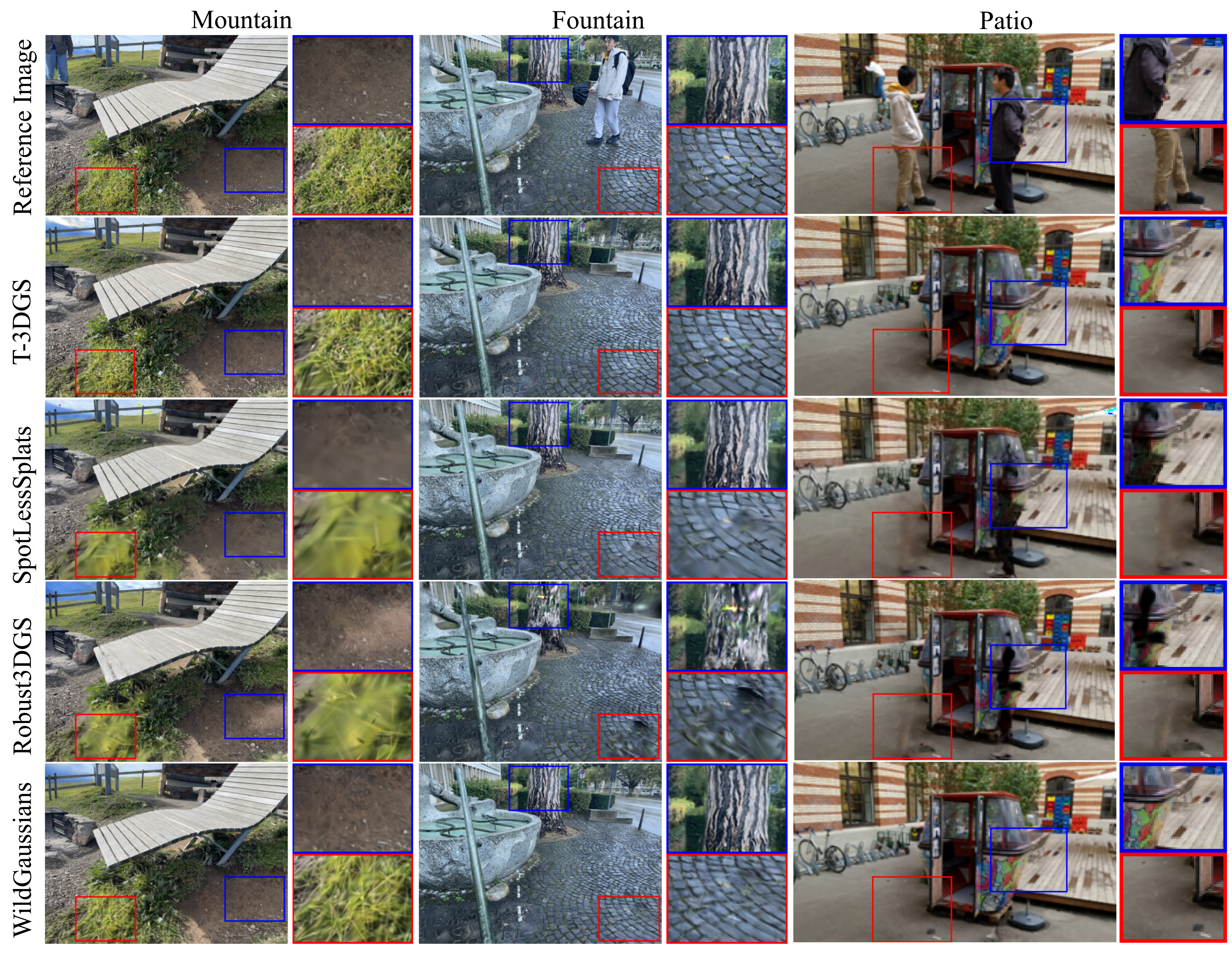}
  \caption{Qualitative results on the \emph{On-the-go} dataset using the training frames. Our method produces higher-quality renderings without artifacts.}
  \label{fig:vis_onthego_renders}
\end{figure*}

\begin{figure*}
  \centering
  \includegraphics[width=\linewidth]{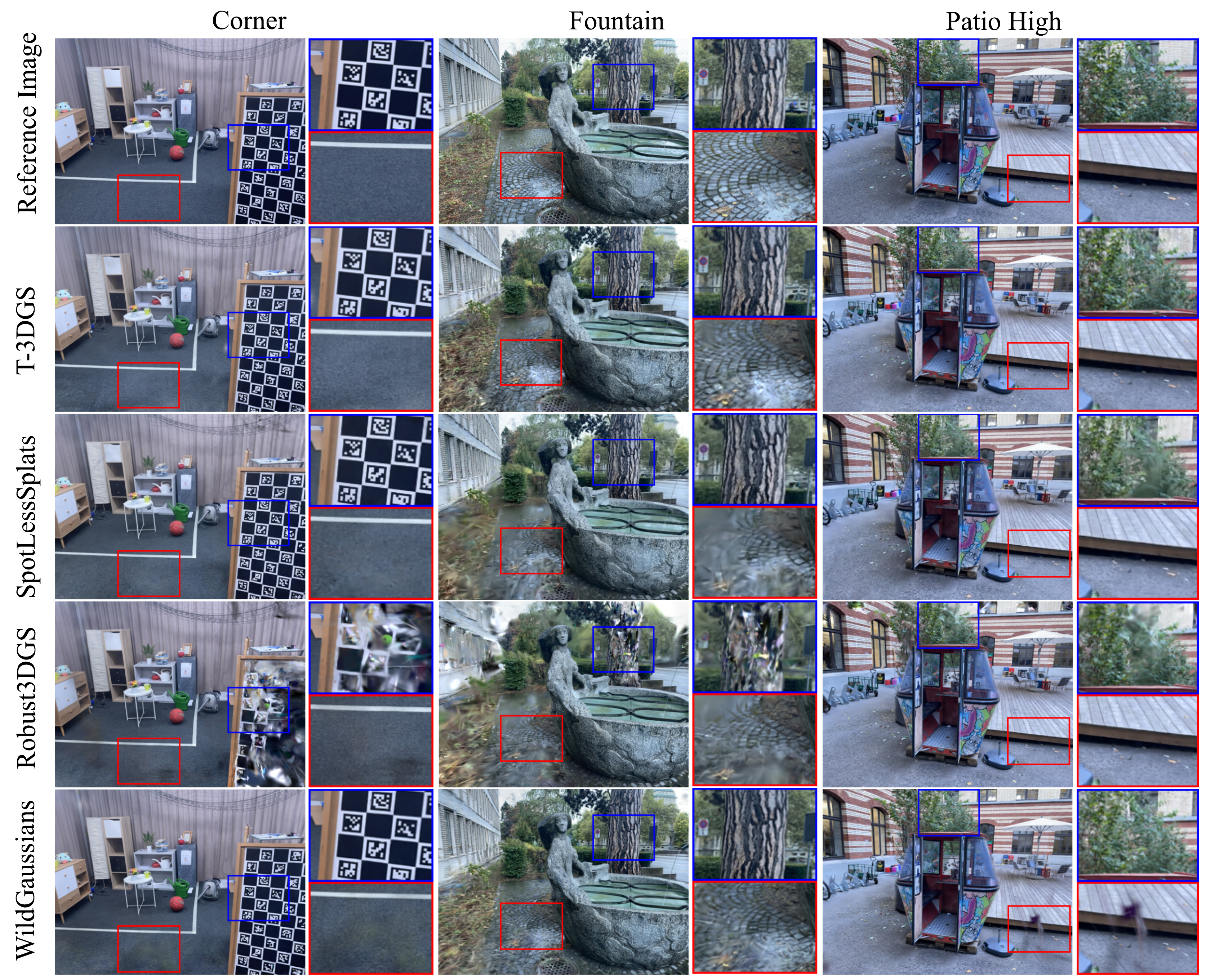}
  \caption{Qualitative results on the \emph{On-the-go} dataset using the testing frames. Our method produces higher-quality renderings without artifacts.}
  \label{fig:vis_onthego_test_renders}
\end{figure*}

\section{Additional Experiments with TMR Module}
Even though our proposed \emph{TMR} module leverages SAM2 to propagate the transient masks, we would like to emphasize that our method enables mask propagation spatially and temporally consistent, thereby providing more accurate and reliable reconstruction.
Table~\ref{tab:mask_comparison} presents an evaluation of the reconstruction quality of WildGaussians with our \emph{TMR} module.
First, we obtained the transient masks using WildGaussians. Then, we propagate them through our \emph{TMR} module. Finally, we reconstruct the scenes based on the transient masks obtained in the previous step.
Our evaluation shows that our method produces higher-quality results for most scenes, with comparable performance in the remaining ones.
Our method, with the \emph{TMR} module, outperforms WildGaussians with the \emph{TMR} module on Anti-Stress, Lab (1), Lab (2) scenes.
The TMR module generally enhances the reconstruction quality of the original WildGaussians, but it is limited because of the false positive transient detections that come from WildGaussians itself.
Furthermore, we note that the hyperparameters of our \emph{TMR} module are highly dependent on the dataset rather than the model. That makes our \emph{TMR} module robust across the different methods.

\section{More Qualitative Comparisons}
For qualitative comparison, we evaluate our method against SpotLessSplats~\cite{sabour2024spotlesssplats}, Robust3DGS~\cite{ungermann2024robust}, and WildGaussians~\cite{kulhanek2024wildgaussians}. We provide corresponding renderings for the masks shown in the main paper in Sec.~4.2. Fig.~\ref{fig:vis_onthego_renders} and~\ref{fig:vis_onthego_test_renders} show reconstructions of several scenes from the \emph{On-the-go} dataset on training and testing frames, respectively. Currently, most methods can produce fairly good reconstructions and avoid significant artifacts, so generally, most methods produce fairly similar results (at least in the absence of semi-transient objects and other adversarial cases). Notably, compared to other residual-based methods, we avoid misclassifying high-frequency details and similar objects.

\begin{figure*}
  \centering
  \includegraphics[width=\linewidth]{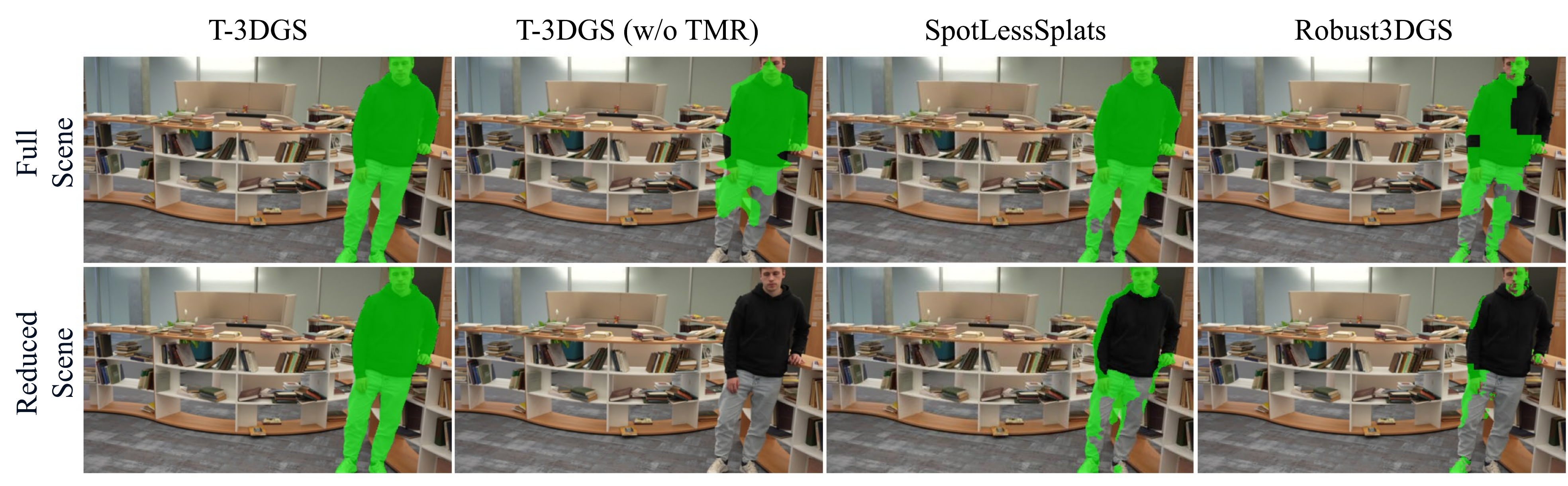}
  \caption{Comparison of predicted masks for full and reduced scenes.}
  \label{fig:full_half_comparison}
\end{figure*}

\begin{figure*}
  \centering
  \includegraphics[width=0.93\linewidth]{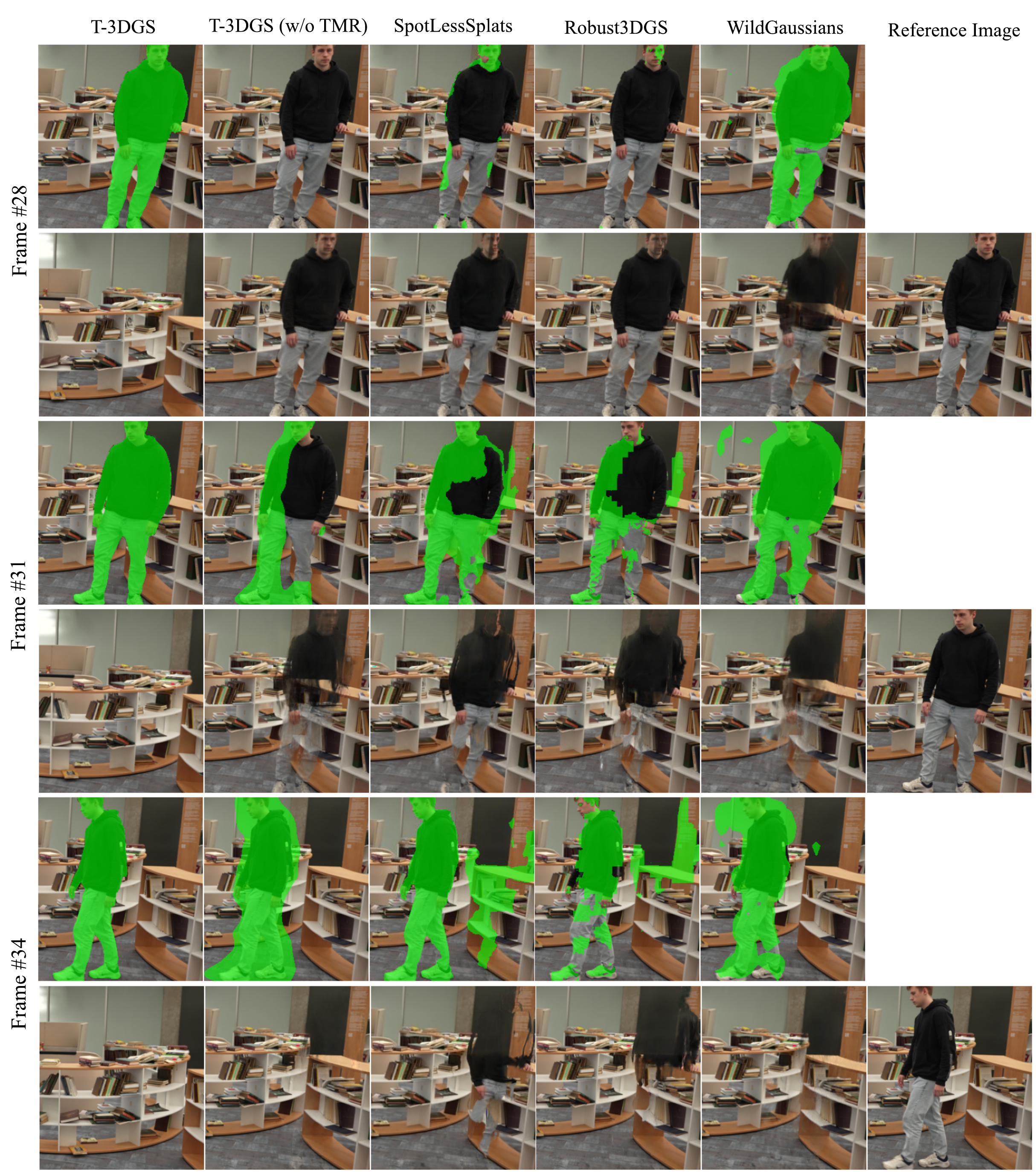}
  \caption{Comparison of predicted masks and scene reconstructions during the movement of semi-transient objects across different frames.}
  \label{fig:consequent_frames_comparison}
\end{figure*}

\section{Handling Semi-Transient Objects}
Semi-transient objects have not been properly addressed in 3D scene reconstruction. Our method represents a significant improvement over previous work and can handle relatively complex scenarios. We provide details on the data capture process and the methodology employed for handling semi-transient objects. We also discuss the importance of both divergence estimation and mask propagation in handling semi-transient objects. Additionally, we discuss the limitations of our proposed method.

Our dataset includes two versions of some scenes: reduced and full. In reduced scenes, the camera operator moves from one end of the scene to the other. In full scenes, however, the operator retraces this path back to the starting position while semi-transient objects continue to move. As illustrated in Fig.~\ref{fig:full_half_comparison}, our proposed TMR module is essential for achieving good results in reduced scenes, which are particularly challenging. In full scenes, the additional frames lead to significantly improved mask predictions for all models because transient objects remain visible for longer periods. When fewer frames capture the scene, many methods mistakenly classify these transient objects as static. Overall, our findings highlight that effectively handling semi-transient objects is a major challenge in in-the-wild video processing. To develop the most challenging datasets and to rigorously compare different methods, it is important to consider not only the types of motion dynamic objects exhibit but also their movement relative to the camera.

As mentioned above, some of the scenes include semi-transient objects occluding the static scene for prolonged periods of time while remaining mostly still. As this period of time increases, semi-transient objects can effectively become static. Although this effect might seem irrelevant to the detection of dynamic objects, this is not the case. As shown in the Fig.~\ref{fig:consequent_frames_comparison}, most methods mask the static background as if it were masking the semi-transient object. Notably, because WildGaussians relies heavily on semantic information, it can "propagate" the masks. However, this happens too late into the training process while our method avoids this problem, and this highlights the importance of using both divergence estimation and mask propagation algorithm we have proposed. Moreover, we aim to minimize false classifications of static objects as dynamic. As discussed earlier, even WildGaussians produces an excessive number of misclassifications for \emph{TMR}. Therefore, our method is crucial for mask propagation to avoid introducing additional errors. This is in stark contrast to the competing methods, which have a lot more false positives. Mask propagation could introduce additional errors and might not contribute to overall quality improvement.

Our method reliably removes transient and semi-transient distractors and successfully reconstructs static artifact-free 3D scenes. However, we have observed that predicted masks tend to be inflated due to the low resolution of the extracted feature maps. Our method can also produce inconsistent results for small objects, as DINOv2 features are computed on patches. These problems could be addressed by using feature extractors with higher-resolution feature maps or guided upsampling.
Additionally, the temporal refinement process is limited by a memory window of $N_m$ frames, which means that if an object disappears for more than $N_m$ frames and then reappears, it will be treated as a new instance with a different label. This can lead to inconsistent tracking and potentially affect the filtering process, especially for semi-transient objects that may temporarily leave the scene. Furthermore, our current filtering approach using global coverage scores may incorrectly filter out valid dynamic objects that undergo significant size changes, such as objects moving towards or away from the camera, or those experiencing perspective changes. We leave this aspect for future work.

\end{document}